\documentclass[conference]{IEEEtran}
\usepackage{cite}
\usepackage{graphicx}
\usepackage{amsmath,amssymb,amsfonts}
\usepackage{algorithmic}
\usepackage{array}
\usepackage{hyperref}
\usepackage{float}

\title{Real-Time Sign Language Gestures to Speech Transcription using Deep Learning }

\author{Brandone Fonya, Clarence Worrell\\
        \texttt{\{bfonya, cworrell\}@andrew.cmu.edu}\\Carnegie Mellon University}
        
\begin{document}

\maketitle

\begin{abstract}
Communication barriers pose significant challenges for individuals with hearing and speech impairments, often limiting their ability to effectively interact in everyday environments. This project introduces a real-time assistive technology solution that leverages advanced deep learning techniques to translate sign language gestures into textual and audible speech. By employing convolution neural networks (CNN) trained on the Sign Language MNIST dataset, the system accurately classifies hand gestures captured live via webcam. Detected gestures are instantaneously translated into their corresponding meanings and transcribed into spoken language using text-to-speech synthesis, thus facilitating seamless communication. Comprehensive experiments demonstrate high model accuracy and robust real-time performance with some latency, highlighting the system’s practical applicability as an accessible, reliable, and user-friendly tool for enhancing the autonomy and integration of sign language users in diverse social settings.
\end{abstract}

\begin{IEEEkeywords}
Sign language classification, sign to speech, convolution neural network, deep learning.
\end{IEEEkeywords}

\section{Introduction}
Communication remains a fundamental human need, yet individuals with visual impairments face significant barriers when attempting to understand sign language, which is inherently visual. While sign language serves as a primary mode of communication for many people in the deaf and hard-of-hearing community, it offers limited accessibility for visually impaired individuals or those unfamiliar with sign gestures. Bridging this communication gap is essential for promoting inclusive and equitable access to information.

This project addresses the challenge by designing and implementing a real-time assistive system that leverages deep learning techniques to translate static sign language gestures into both textual and auditory outputs. Specifically, the system employs a Convolution Neural Network (CNN) trained on the Sign Language MNIST dataset to classify American Sign Language (ASL) ~\cite{ASL} hand signs from live video input. The classified gestures are then converted into spoken words using a text-to-speech engine, enabling real-time audio feedback.
The proposed system was developed using Python and integrates several key libraries and frameworks. OpenCV facilitates video capture from a webcam, MediaPipe is used for hand detection and landmark tracking, and pyttsx3 library for text to speech conversion in python. provides offline speech synthesis capabilities. Together, these components form a pipeline capable of capturing, classifying, and vocalizing hand gestures in real time using only a standard computing device and webcam.
By combining deep learning with computer vision and speech technology, this work contributes to the development of low-cost, accessible solutions that support inclusive communication for visually impaired individuals and foster greater interaction between signers and non-signers.

\section{Problem Statement}
According to the World Health Organization (WHO), over 2.2 billion people globally have a near or distance vision impairment, with at least 1 billion cases preventable or unaddressed ~\cite{WHO}. Among these individuals, many face significant barriers in accessing and processing visual information, including written language. This limitation severely restricts their ability to communicate in environments where visual cues are dominant.
Sign language, used by over 70 million deaf individuals worldwide ~\cite{who2025report}, remains a key method of communication. However, its use requires both sender and receiver to understand the visual gestures, rendering it ineffective for the visually impaired or those unfamiliar with sign language.
While various assistive technologies exist for text-to-speech conversion, there remains a gap in real-time interpretation of sign language gestures into audible speech. Bridging this gap would significantly enhance inclusivity by enabling seamless communication between deaf individuals using sign language and blind or sighted individuals unfamiliar with it.
Moreover, existing research and commercial solutions often rely on expensive hardware such as multi-camera setups or specialized gloves. These systems are inaccessible to users in low-resource settings and lack portability, making them impractical for everyday use.

\subsection{Key challenges}

\begin{itemize}
    \item Dynamic Gesture Recognition: Classifying sign language gestures accurately in real time from a continuous video stream is computationally intensive and sensitive to lighting, background clutter, and hand orientation.
    \item Robustness to Variation: Variability in hand sizes, skin tones, and gestures must be handled reliably across diverse users.
    \item Accessibility and Affordability: Solutions should be lightweight, deployable on standard laptops with webcams, and usable without specialized hardware.
\end{itemize}

\subsection{ Project Objective}
To address this problem, this project aims to develop a real-time assistive system that:

1. Capture and classify American Sign Language (ASL) gestures using a deep learning model trained on the Sign Language MNIST dataset.

2. Convert classified gestures into readable text and audible speech output using a text-to-speech engine, running in real time using only a webcam and standard hardware, providing an affordable and accessible solution for individuals with visual impairments.

\section{Literature Review}
Sign language recognition systems have gained increasing importance for facilitating communication between Deaf and hearing communities with recent studies demonstrating their transformative potential in healthcare~\cite{importance}.

Current sign language recognition systems achieve notable performance benchmarks through various approaches. Transformer and LSTM-based architectures now reaches 88\% accuracy on continuous sign recognition~\cite{lstm-based}. However, these models still face challenges when deployed in real-world environments, where accuracy can drop below 68\% due to lighting variations and other factors~\cite{lighting}.

Gesture-to-speech conversion introduces additional technical hurdles. Current systems must address coarticulation effects in continuous signing, speaker-dependent variations in signing style, and precise audio-visual synchronization. Graph-based methods show promise for improved articulation modeling~\cite{graph-method}, but state of the art (SOTA) end-to-end systems still achieve moderate speech transcription accuracy~\cite{transcription}. This suggests significant opportunities for improvement in both recognition accuracy and translation quality.
 
\section{Research Questions}
1. "How effectively can a deep learning model, use convolution neural networks (CNNs) to translate dynamic sign language gestures into text"

2. “How can I use text to speech, convert predicted sign gestures into audible speech for visually impaired in real time”

\section{Methodology}
The project was implemented in two main stages: model development and real-time application integration.
In the first stage, a Python script was developed using TensorFlow and Keras to train a CNN model for image-based sign language classification. The model was trained on the Sign Language MNIST dataset, which comprises 28×28-pixel grayscale images of static hand gestures representing 24 alphabet letters (A–Y, excluding J and Z). Each image is labelled with its corresponding letter, and the dataset includes approximately 27,000 training samples and 7,000 test samples, stored in CSV format.
In the second stage, a real-time classification program was built using OpenCV, MediaPipe, and pyttsx3. This application captures hand gestures from a live webcam feed, detects the hand region, and preprocesses the extracted image frame to match the input format expected by the trained CNN model. The model then classifies the gesture, and the predicted output is converted into audible speech using a text-to-speech engine. This complete pipeline enables the system to translate sign language gestures into spoken words in real time, enhancing communication accessibility for visually impaired individuals.

\subsection{Dataset}
The dataset used was the Sign Language MNIST dataset found on Kaggle, which consists of 28x28 grayscale images of hand signs representing 24 letters of the alphabet (A to Y, excluding J and Z). It includes approximately 27,000 training samples and 7,000 test samples, stored in CSV format with labels and pixel values.

\begin{figure}[H]
    \centering
    \includegraphics[width=0.5\textwidth]{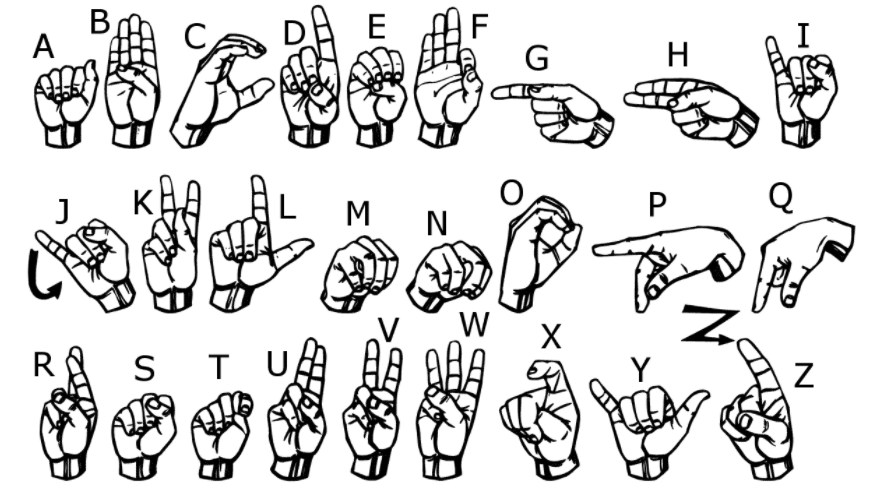}
    \caption{Hand signs in the American sign language MNIST dataset}
    \label{fig:dataset-image}
\end{figure}

The model development approach and implementation in a real time sign language detection is explained below.

\subsection{Part 1: Model Development}
\subsubsection{Load and processed the data}
To prepare the data for model training, a load data function was created to load the Sign Language MNIST train and test datasets, where each row consists of a label and 784 pixel values representing a 28×28 grayscale image. To ensure label consistency, invalid entries were filtered out, retaining only those with values from 0 to 23. This filtering, implemented using isin(range(24)), ensured compatibility with the sparse categorical crossentropy loss function, which requires integer labels within that range.
After filtering, the labels for training and testing were extracted, and the pixel values were normalized by dividing by 255, scaling them to the [0, 1] range. This normalization step improves gradient stability and accelerates convergence during training. The flattened pixel arrays were then reshaped into 28×28×1 tensors to match the expected input shape for the model’s Conv2D layers, which require height, width, and channel dimensions. The function returns the processed and structured training and test data, ready for input into the CNN model.

\subsubsection{Building the CNN model}
A build model function was developed to construct the Convolutional Neural Network (CNN) using Keras's Sequential architecture. This approach allows layers to be stacked in a linear fashion, ideal for image classification tasks that do not require complex or branching connections.
The first layer added was a convolutional layer with 32 filters of size 3×3, designed to detect low-level features such as edges and corners in the 28×28 grayscale input images. The ReLU activation function was applied to introduce non-linearity and enhance learning capacity.
To reduce computational complexity and control overfitting, a MaxPooling2D layer with a 2×2 window followed the first convolution, effectively down sampling the feature map to a size of approximately 13×13 while retaining key spatial features.
A second convolutional block was then introduced, this time with 64 filters, enabling the network to capture more abstract patterns like hand shapes. Another 2×2 max pooling layer followed to reduce the output further to around 5×5. This layer also used the ReLU activation function to maintain consistency and non-linearity.
A third convolutional layer with 128 filters was added to extract higher-level features unique to specific hand gestures, improving the model’s ability to differentiate between similar signs.
After the convolutional layers, a Flatten layer was used to transform the feature maps into a one-dimensional vector, preparing the data for the fully connected layers. A Dense layer with 256 units and ReLU activation was added to learn complex combinations of features. To reduce the risk of overfitting, a Dropout layer with a rate of 0.5 was used, which randomly disables 50\% of the neurons during training. This was particularly effective given the dataset's size of approximately 27,000 training samples.
The final output layer consisted of a Dense layer with 24 units (corresponding to the 24 alphabet classes) and used a Softmax activation function to produce a probability distribution over all possible gesture classes.

\subsubsection{Training the model}
Created a train model function to handle the training process of the CNN model. This function accepts the pre-processed training data (features and labels), uses the previously defined build model function to construct the architecture, and then compiles the model in preparation for training.
During compilation, the Adam optimizer was selected for its adaptive learning rate and efficient convergence properties, making it suitable for training deep learning models. The loss function used was sparse categorical cross entropy, which is ideal for multi-class classification tasks with integer-labeled targets ranging from 0 to 23, as is the case with the Sign Language MNIST dataset.
To evaluate the model's performance during training, accuracy was chosen as the primary metric. This metric offers a clear view of how well the model is predicting the correct gesture classes.
To avoid overfitting and optimize training time, an early stopping mechanism was implemented. Training was automatically halted if the validation loss did not improve for five consecutive epochs, and the model was configured to restore the best weights from the
epoch with the lowest validation loss. This ensures that the final model performs optimally on unseen data.
The model was trained for up to 30 epochs with a batch size of 64. Additionally, 20\% of the training data was set aside for validation to monitor the model's generalization performance during training.
After training was complete, the model was saved in Keras format (.keras). This format supports easy reloading for future use and enabled seamless integration into the real-time sign language classification program developed in the second part of the project.

\subsubsection{Evaluating model performance}
Evaluated the model’s performance using the test dataset to compute its loss and accuracy. Plotted the train vs validation accuracy and train vs validation loss

\subsection{Part 2: Real-Time Application Integration}
\subsubsection{Building real time classifier python program which uses the trained model}

Created a RealTimeClassifier python class, defined to perform real-time sign language classification using the pre-trained Keras model, with initialization of the model, MediaPipe Hands for hand detection, and pyttsx3 for text-to-speech functionality, along with a label mapping for 25 letters (A-Y, excluding Z).

In the preprocess method, an input image is converted to grayscale, resized to 28x28 pixels, reshaped to (1, 28, 28, 1) to match the model’s input shape, and normalized by dividing pixel values by 255.0.
The predict method processes the image using the preprocess method, feeds it into the model to get predictions, extracts the class with the highest probability using np.argmax, and returns the corresponding letter label and confidence score.
The text to speech function created, uses pyttsx3 to convert a predicted label into spoken text, playing the audio output for the classified letter.

In the classify from camera method, a video capture was initiated using OpenCV to access the webcam, continuously reading frames until the user presses 'q' to quit.
Converted each frame to RGB, processed by MediaPipe Hands to detect hand landmarks, and the bounding box coordinates of the detected hand are calculated with a 20-pixel padding.
Extracted the hand region from the frame using the bounding box, preprocessed, and passed to the predict method to classify the hand sign, returning the predicted letter and confidence.

A green rectangle is drawn around the detected hand, and the predicted label with confidence is displayed on the frame in red text above the rectangle using OpenCV.
If the prediction confidence exceeds 0.8, the text to speech method is called to vocalize the predicted letter (with high confidence), providing audio feedback.
The frame with annotations is displayed in a window titled 'Real-Time Sign Language Classification', and the loop continues until the user presses 'q', at which point the video capture is released and the window is closed.
The main block creates an instance of RealTimeClassifier and calls classify from camera to start the real-time classification process.

\section{Results}
The final trained model achieved a test accuracy of 95.72\% and a test loss of 0.2106 on the Sign Language MNIST dataset. With a total of 394,008 trainable parameters, the model demonstrated a balanced complexity, sufficient to capture meaningful patterns while minimizing overfitting risks through dropout regularization.
In real-time testing, the model correctly identified hand gestures with high confidence and the corresponding text-to-speech feedback functioned reliably. However, the program exhibited noticeable latency due to the hand detection frequency in MediaPipe, which could be optimized.
The image below shows the training history of the model from epoch 1 to 30, showing the changes in the loss and accuracies over the train and validation dataset.

Total Parameters: 1,182,026.
This is the total number of parameters in the model, including both trainable and optimizer-related parameters.
Adding up the trainable parameters from each layer:

\begin{table}[H]
\centering
\caption{Trainable Parameters in Model Architecture}
\label{tab:model-summary}
\begin{tabular}{|l|r|}
\hline
\textbf{Layer}       & \textbf{Trainable Parameters} \\
\hline
Conv2d               & 320                           \\
Max\_pooling         & 0                             \\
Conv2d               & 18,496                        \\
Max\_pooling         & 0                             \\
Conv2d               & 73,856                        \\
flatten              & 0                             \\
dense                & 295,168                       \\
dropout              & 0                             \\
dense                & 6,168                         \\
\hline
\end{tabular}
\end{table}

Table~\ref{tab:model-summary} summarizes the trainable parameters across different layers of the convolutional neural network used in the model.

Total = 320 + 18,496 + 73,856 + 295,168 + 6,168 = 394,008 trainable parameters.
These are the parameters the model learned during training from each layer, such as weights and biases in the convolutional and dense layers. With 394,008 trainable parameters, the model had sufficient capacity to learn the patterns in the Sign MNIST, but it’s not excessively large, reducing the risk of overfitting, especially with dropout regularization in place.

\begin{table}[H]
\centering
\caption{Model Performance on Test dataset}
\label{tab:test-summary}
\begin{tabular}{|l|r|}
\hline
\textbf{Metric} & \textbf{Value} \\
\hline
Test Accuracy & 0.95716 \\
Test Loss & 0.21059 \\
\hline
Precision (Macro Avg) & 0.96 \\
Recall (Macro Avg) & 0.95 \\
F1-Score (Macro Avg) & 0.95 \\
\hline
\end{tabular}
\end{table}

\begin{figure}[H]
    \centering
    \includegraphics[width=0.5\textwidth]{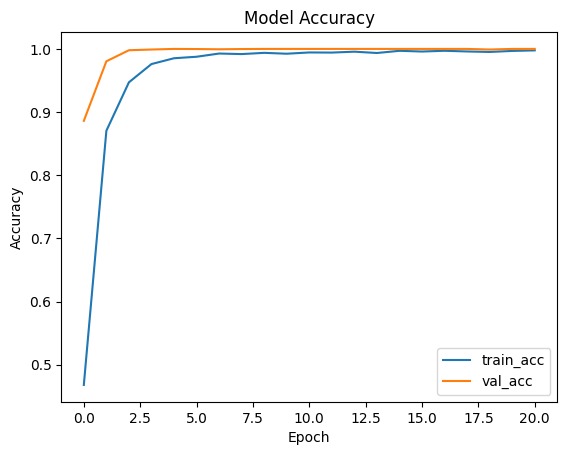}
    \caption{Model training and validation accuracy over epochs. The model demonstrates rapid convergence with validation accuracy stabilizing above 99\% after only a few epochs.}
    \label{fig:model-accuracy}
\end{figure}

\begin{figure}[H]
    \centering
    \includegraphics[width=0.5\textwidth]{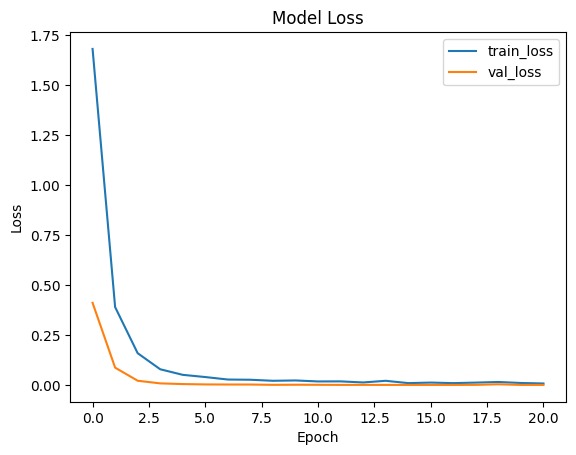}
    \caption{Model training and validation loss over epochs.}
    \label{fig:model-accuracy}
\end{figure}

\subsection{Model interpretation}
The trained model demonstrated strong performance on the Sign Language MNIST dataset, achieving a high test accuracy of 95.72\% and a low test loss of 0.2106. These results indicate the model's effectiveness in accurately classifying 24 static hand sign letters (A to Y, excluding J and Z), and its capacity to generalize well to unseen test data. The macro-averaged precision (0.96), recall (0.95), and F1-score (0.95) all indicate balanced performance across most classes. The tight alignment between precision and recall scores also shows the models ability with relatively few false positives or negatives.

From the training curves, the training accuracy began around 0.5 and rapidly increased to 1.0 by epoch 5, showing that the model quickly learned the patterns in the training data. The validation accuracy rose to approximately 0.95 by epoch 3 and remained stable thereafter, indicating strong generalization with minimal signs of overfitting.

Similarly, the training loss dropped steeply from 1.75 to below 0.25 within the first five epochs, while the validation loss mirrored this trend, falling from 0.5 to around 0.2. These consistent trends between training and validation losses further reinforce the model’s ability to converge effectively without overfitting.

After epoch 5, both accuracy and loss curves plateaued, and the model's performance metrics remained close throughout the remaining training epochs. This behavior confirms the effectiveness of the early stopping mechanism (configured with a patience of 5), which successfully halted training once the validation loss showed no further improvement, preserving the model's best weights.

Although there was a slight performance gap between training and validation metrics—1.0 vs. 0.95 in accuracy and 0.1 vs. 0.2 in loss—this is within acceptable limits and suggests only a minor degree of overfitting. Overall, the model’s high test accuracy confirms its reliability for real-world application in sign language classification tasks. The small gap between training and validation metrics suggests minimal overfitting and strong generalization of the model.

\subsection{Testing}
\begin{figure}[H]
    \centering
    \includegraphics[width=0.5\textwidth]{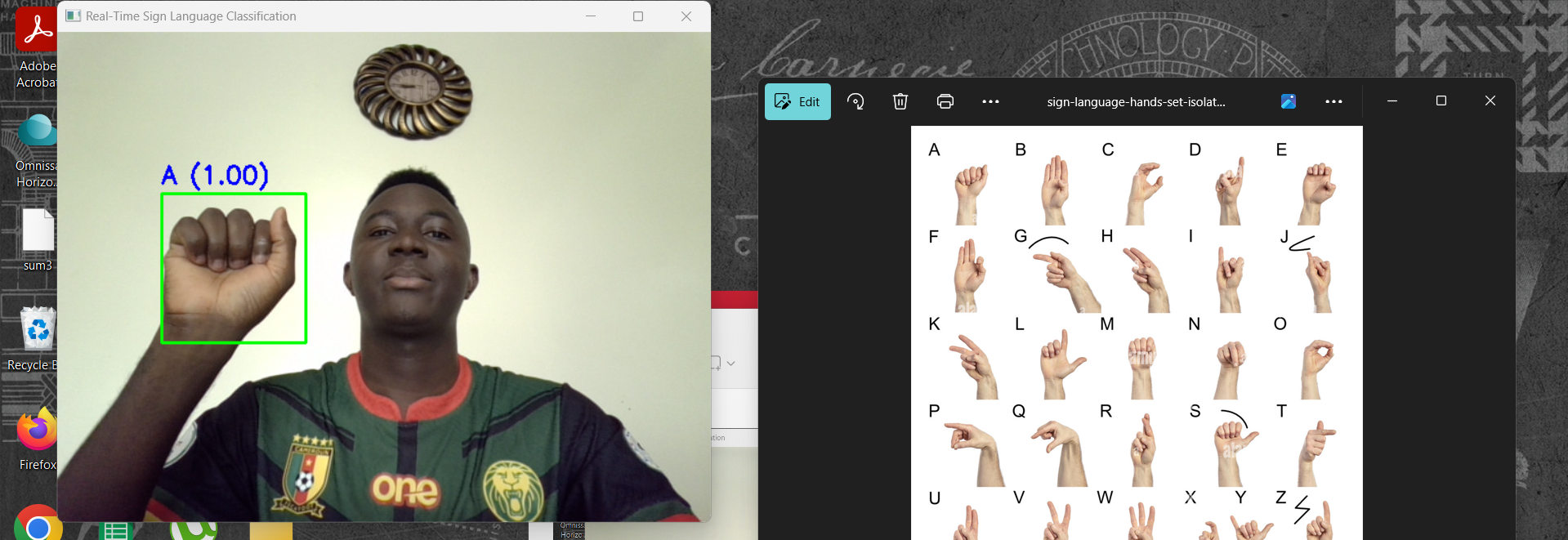}
    \caption{Image showing program running in real time, performaning sign language gesture detection with confidence level shown.}
    \label{fig:program-test}
\end{figure}
The image in ~\ref{fig:program-test} above shows a demo screenshot of a real-time program testing with an ‘A’ hand gesture with (Accuracy/confidence level of 1.00). When the program starts running, it continuously captures the images frame by frame. When the hand is detected, it classifies it using the build CNN model.

\section{Conclusion}
Leveraging a CNN model trained on the Sign Language MNIST dataset, the model achieved a high test accuracy of 95.72\%, successfully developed and deployed a real-time assistive system that converts static sign language gestures into both text and audible speech, thereby enhancing communication accessibility for visually impaired individuals

The integration of real-time video processing with MediaPipe for hand detection, OpenCV for frame manipulation, and text to speech synthesis allowed translation of hand gestures into spoken language. The system runs effectively on standard hardware, requiring only a webcam, making it a low-cost and accessible solution.

While the current version supports only static gestures, it provides a strong foundation for future expansion to dynamic signs and continuous gesture recognition. Overall, this project contributes meaningfully toward inclusive AI solutions by empowering individuals with visual impairments to interact with sign language users in a more seamless and accessible way utilizing the power of deep learning.

\section{Future Work}
This research opens several promising directions for future investigation. Firstly, to extend the pipeline to support a wider variety of sign language datasets, including non-Western sign languages like Japanese Sign Language (JSL) and Indian Sign Language (ISL), to improve cultural inclusivity. Secondly, to develop more robust methods for capturing and interpreting long, complex sentences from continuous sign gestures, which remains a significant challenge in real-world applications. Thirdly, to investigate techniques which could reduce the latency in speech transcription systems while maintaining accuracy, potentially through novel attention mechanisms or hybrid architectures.

\section*{Acknowledgment}
Sincere gratitude to Professor Clarence Worrell for permitting me conduct this research as part of his Applied Deep Learning course at Carnegie Mellon University. His guidance and insights were invaluable throughout this project. I also extend my sincere gratitude to the teaching assistant and course mates who provided technical support, constructive feedback and reviews during the project phases.

\section*{Conflict of Interest}
The author declares that there is no conflict of interest regarding the publication of this work. No financial or personal relationships exist that could inappropriately influence or bias the content of this work. All data and materials used in this research were obtained and processed as open source in accordance with appropriate ethical guidelines.

\end{document}